\documentclass{article}
\usepackage{arxiv}
\usepackage[utf8]{inputenc} 
\usepackage[T1]{fontenc}    
\usepackage{hyperref}       
\usepackage{url}            
\usepackage{booktabs}       
\usepackage{amsfonts}       
\usepackage{nicefrac}       
\usepackage{microtype}      
\usepackage{lipsum}		
\usepackage{graphicx}
\usepackage{natbib}
\usepackage{doi}
\usepackage{hyperref}
\usepackage{graphicx}
\usepackage{amsmath}
\usepackage{multirow}
\usepackage[table,xcdraw]{xcolor}
\usepackage{graphicx}
\usepackage{subcaption}
\usepackage{amsmath,amssymb,amsfonts}
\usepackage{bm}

\hypersetup{
    colorlinks=true,     
    linkcolor=black,     
    urlcolor=black,      
    citecolor=blue,
}

\title{A MIL Approach for Anomaly Detection in Surveillance Videos from Multiple Camera Views}
\date{} 
\author{ \href{https://orcid.org/0000-0003-2535-3024}{\includegraphics[scale=0.06]{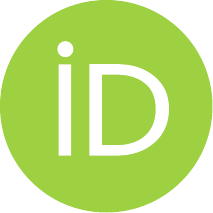}\hspace{1mm}Silas Santiago. Lopes ~Pereira} \\
	Federal Institute of Education, Science, and Technology of Ceará (IFCE) - Aracati, CE, Brazil~ \\
 State University of Ceará -- CCT-PPGCC-UECE -- 60740-903 -- Fortaleza, CE -- Brazil~\\
	\texttt{silas.santiago@ifce.edu.br} \\
	\And
	\href{https://orcid.org/0000-0002-4983-1724}{\includegraphics[scale=0.06]{orcid.pdf}\hspace{1mm}Jos\'e Everardo Bessa ~Maia} \\
        State University of Ceará -- CCT-PPGCC-UECE -- 60740-903 -- Fortaleza, CE -- Brazil~\\
	\texttt{jose.maia@uece.br} \\
}

\renewcommand{\undertitle}{Technical Report}
\hypersetup{
pdftitle={A MIL Approach for Anomaly Detection in Surveillance Videos from Multiple Camera Views},
pdfsubject={q-bio.NC, q-bio.QM},
pdfauthor={David S.~Hippocampus, Elias D.~Striatum},
pdfkeywords={First keyword, Second keyword, More},
}

\begin{document}
\maketitle

\begin{abstract}
Occlusion and clutter are two scene states that make it difficult to detect anomalies in surveillance video. Furthermore, anomaly events are rare and, as a consequence, class imbalance and lack of labeled anomaly data are also key features of this task. Therefore, weakly supervised methods are heavily researched for this application. In this paper, we tackle these typical problems of anomaly detection in surveillance video by combining Multiple Instance Learning (MIL) to deal with the lack of labels and Multiple Camera Views (MC) to reduce occlusion and clutter effects. In the resulting MC-MIL algorithm we apply a multiple camera combined loss function to train a regression network with Sultani’s MIL ranking function. To evaluate the MC-MIL algorithm first proposed here, the multiple camera PETS-2009 benchmark dataset was re-labeled for the anomaly detection task from multiple camera views. The result shows a significant performance improvement in F1 score compared to the single-camera configuration.
\end{abstract}


\section{Introduction}

In video surveillance scenarios for detecting anomalous events, the use of a single camera view to identifying suspicious activities and abnormal behavior brings with it a set of limitations and difficulties for automating this task. In addition to variability factors such as lighting conditions, background clutter, and low viewing resolution, the information captured is also dependent on the proper calibration of the camera for the target environment. This dependence on perspective can often generate uncertainty regarding the interpretation of actions and behaviors. Another limitation is the occlusion of people or objects, which can impair the recognition of anomalous activity. In this sense, the use of multiple overlapping cameras to capture and monitor the same scene can provide a general perspective of the whole scenario,  greater representation of information and a greater amount of data from multiple perspectives.

However, video anomaly detection (VAD) is a challenging problem in the computer vision area. The definition of an anomaly involves subjectivity, depends on localization and context, and can vary in duration and content. Thus, the definition of an anomaly could become a complex problem. Beyond that, there is also the existing challenge of capturing anomaly examples. Performing frame and pixel-level annotations could be a tedious and expensive human activity, which leads to the creation of frequently unbalanced databases and the disseminated use of unary classification methods, although the existence of works on literature about binary classification \cite{wan2020weakly}.

Multiple Instance Learning (MIL) is applicable when the knowledge about label categories and training samples is incomplete \cite{AliEtAl2008ActRecogMIL}. In MIL, some bags contain multiple instances instead of individual ones to represent each pattern in a dataset. From a binary perspective, a bag could have normal or anomalous labels, which can be used to train a model with an appropriate machine-learning technique. Different factors can impact the performance of MIL approaches. Firstly, predictions can be executed at the bag or instance levels, and these two levels have distinct misclassification costs. Second, the composition of each bag, such as the proportion of examples from each category and the relation between examples, impacts the performance of MIL methods. Third, ambiguity in instance labels can be related to label noise as well to instances not belonging to classes. Finally, class distributions can also affect MIL algorithms depending on their assumptions about the data (\cite{CarbonneauEtAl2018MILSurvey}). Recent studies have shown the performance efficiency of the MIL approach in detection and recognition tasks.

Although using the weakly supervised approaches mitigates the need for labeled training data, acquiring video data labeled even at the video level is still an exhaustive and challenging task, and anomaly detection task in realistic scenes is still an open problem. The vast majority of  video anomaly detection works explore single-camera approaches and they do not explore the intrinsic information existing in multiple camera views for a same scene. In this sense, since collected video events can be described by multiple overlapped camera perspectives to describe a same scene, is crucial to explore multi-camera strategies to learn the underlying semantics of each camera view of the same scene data. Thus, we propose a multi-camera multiple-instance training scheme in this work. We employ the MIL algorithm of \cite{SultaniEtAl2018} in addition to a combined loss function to take into account the multiple views of the same data during the network weight adjustment. We consider a multi-camera video anomaly detection dataset generated from PETS 2009 dataset for evaluation and comparison with the proposed multi-camera approach with the vanilla single-camera case.
The main contributions of this work are summarized as follows:
\begin{itemize}

\item We have developed a MIL training strategy with multiple camera views which improves its performance over single-camera configuration;
\item To evaluate the proposed approach, we re-label the multiple camera PETS-2009 benchmark dataset for the anomaly detection task from multiple camera views;
\item Since \cite{SultaniEtAl2018}  trains your regression network with video bags with a fixed number of segments of varying length, we provide an adaptation in the source code to enable, instead, training with video bags with a variable number of video clips of the same length.
\end{itemize}

This work is organized as follows: Section 2 describes the dataset and the multi-camera multiple-instance proposed method. In section 3, the experimental results are presented and discussed. 
Section 4 presents some relevant related works associated with video anomaly detection.
Section 5 presents the conclusions and directions for future research.

\section{Data and Methods}

This section describes the main steps for data preparation, modeling, and evaluation of our proposed multi-camera multiple-instance video anomaly detection approach. Firstly, we describe the formation of the multi-camera video anomaly dataset used in our experiments. Then, we explain the modelling and evaluation process.

In our work, we consider the multi-camera video anomaly detection problem in the perspective of a regression problem. We describe this problem as follows: Let $ X = \{x_i \} ^ n_ {i = 1} $ a dataset consisting of $n$ video scenes.  Each video scene $x_i$ is composed of multiple videos corresponding to multiple overlapped camera perspectives for a same scene. Each video $x_i$ has also a duration $ t_i $, so that $ T = \{t_i \} ^ n_ {i = 1} $ is the temporal duration of the dataset. Let $ Y = \{y_i \} ^ n_{i = 1} $ be the binary labels for each video in dataset $ X $. We desire to built a predictive model which receives a given video $x_{test}$ and produces as inference an anomaly score. 

\subsection{The Multiple Camera PETS-2009 Benchmark Dataset Re-labeled}

\begin{figure*}[!ht]
\centering
\begin{tabular}{cccc}
     \includegraphics[width=0.18\linewidth]{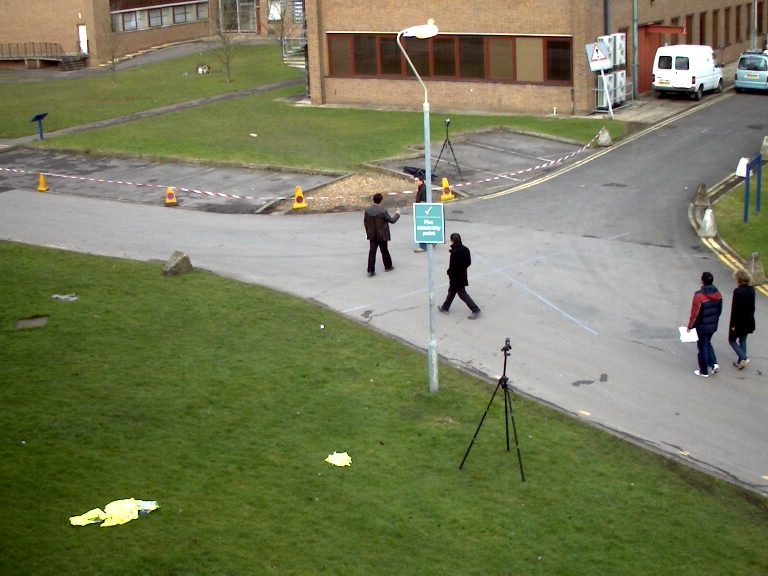} &
     \includegraphics[width=0.18\linewidth]{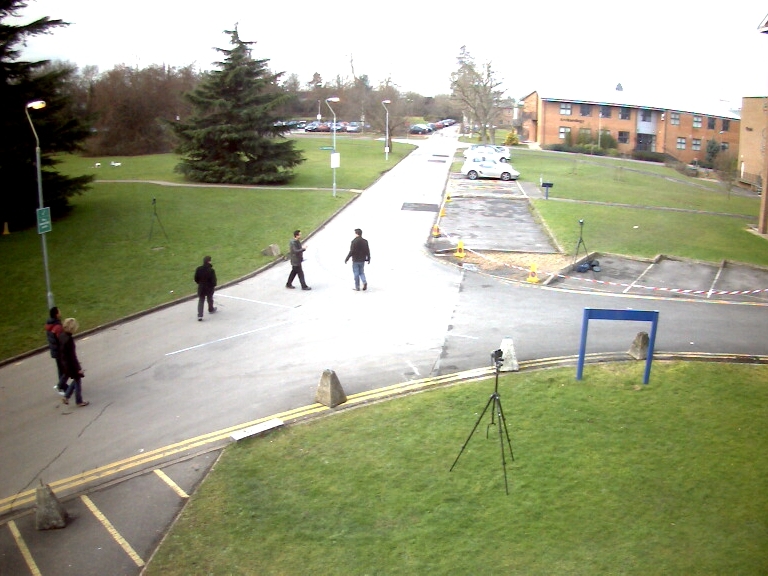} &
     \includegraphics[width=0.18\linewidth]{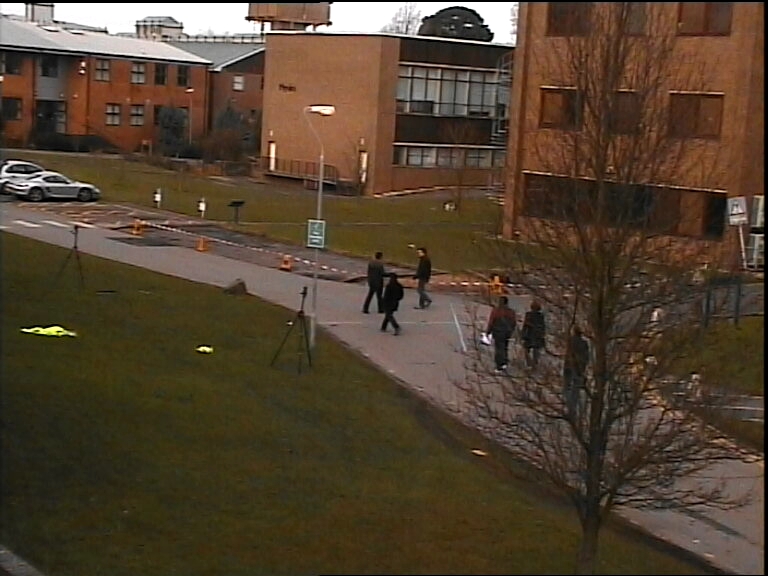} &
     \includegraphics[width=0.18\linewidth]{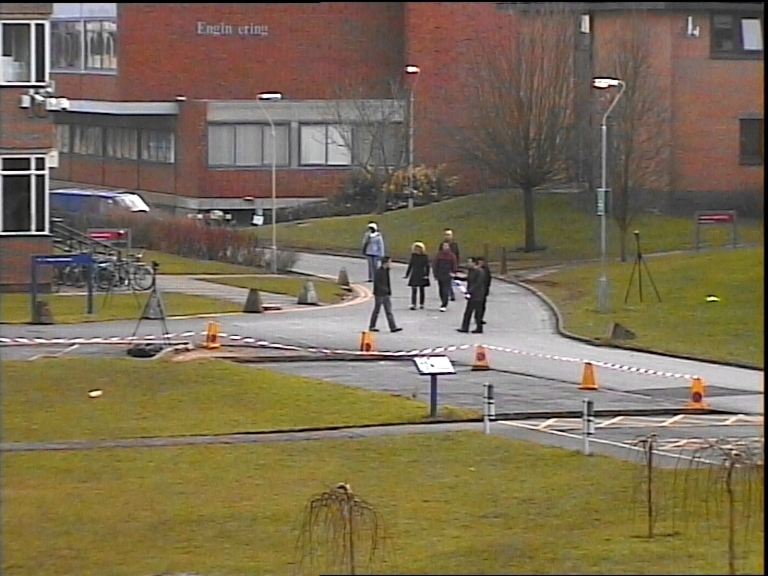} \\
         (a) Camera 1 & (b) Camera 2 & (c) Camera 3 & (d) Camera 4 \\
\end{tabular}
\caption{Multiple Camera Perspectives in \textit{PETS 2009}}
\label{fig:pets2009-4cams}
\end{figure*}

PETS-2009\footnote{\url{https://cs.binghamton.edu/~mrldata/pets2009}} is a benchmark dataset that aggregates different scene sets with multiple overlapped camera views and distinct events involving crowds (\cite{zhang2015camera}). 
We use these frame sequences to derive a new dataset for the multi-camera video anomaly detection task. 
We consider the first four cameras in the original frame sequences, which provide different overlapped visions of the same scene from varying positions and angles, as illustrated in Figure \ref{fig:pets2009-4cams}.

The scenes were labeled at \textit{frame} level as anomaly or normal events. Scenes with \textit{background}, people walking individually or in a crowd, and regular passing of cars are considered as normal patterns. \textit{Frames} with occurrences of people running (individually or in crowd), crowding of people in the middle of the traffic intersection, and people in the counterflow were considered anomalous patterns. 
In summary, there are $27$ scenes, where $19$ scenes reflect normal events while $8$ scenes have some anomalous activity. In these $27$ scenes, there is a total of $528$ clips for each one of the four viewpoints.
In Table \ref{tab:class-distr}, we summarize the distribution of normal and anomalous patterns in the video anomaly detection dataset. 

\begin{table}[!ht]
    \centering
    \begin{tabular}{|l|c|c|c|}
         \hline
         \textbf{Category} & \textbf{Scenes} & \textbf{Clips per Camera} & \textbf{Frames per Camera} \\
         \hline
         Normal & 19 & 409 & 6544 \\
         Anomaly & 8 & 119 & 1904 \\
         \hline
         Total & 27 & 528 & 8448 \\
         \hline
    \end{tabular}
    \vspace{0.2cm}
    \caption{Class Distribution of the Video Anomaly Detection Dataset}
    \label{tab:class-distr}
\end{table}

Since the number of frames is sometimes different among the four cameras of the same scene in some of the videos, we complete each frame sequence with background frames so that the four camera views  have the same number of frames, and the number of frames is a multiple of $16$.
A set of RGB I3D (Inflated 3D) attributes were obtained for each sequence of $16$-frame video clips in the videos. For this step, we use the  \textit{Video Features} library
\footnote{\url{https://github.com/v-iashin/video_features}} that uses a pre-trained model on the \textit{Kinetics 400} dataset. 
We describe the composition of training and test splits for modeling and performance evaluation as follows. Initially, we load the dataset $\mathbf{D} = \{(X_i, y_i, {yf}_{i})\}_{i=1}^N$ with the processed videos scenes. 
Then, we  partitioned $ \mathbf{D} $ into training and test datasets, and we used 50\% of data for further training with a holdout procedure. 
We build both partitions so that we maintain the same proportion of anomalous and normal instances in training and test partitions. The training split contains $9$ normal videos and $4$ anomalous ones for each camera. The test set contains $10$ normal videos and $4$ anomalous videos for each camera.
We made this processed dataset available at the following link: \url{https://github.com/santiagosilas/MC-VAD-Dataset-BasedOn-PETS2009}.

\subsection{Modeling and Evaluation}

The Multiple Instance Learning (MIL) problems in the binary classification context can be formally specified as follows: Consider an instance space $X=\mathbb{R}^d$ and a set of labels $Y= \{0, 1\}$. A model is then built from a dataset with $m$ bags $\beta = \{\beta_1, \beta_2, \ldots, \beta_m\}$. Each bag $\beta_i = \{ \textbf{x}_{i1}, \ldots, \textbf{x}_{ij}, \ldots, \textbf{x}_{in_i}  \}$ is a set with $n_i$ instances and $\textbf{x}_{ij} \in X$.
During the training step, each bag $\beta_i$ has only the information about the associated bag label $\textbf{y}_i \in Y$, but instance labels are unknown.  The learning goal is to predict the label of an unseen bag and also predict the label of its instances (\cite{zhang2021non}).

\nocite{SultaniEtAl2018}
In \cite{SultaniEtAl2018}, video anomaly detection was treated as a regression problem under MIL in which features are mapped to anomaly scores by the use of a 3-layer fully connected neural network. 
The authors utilized a coarse-grained approach in which videos are divided into a fixed number of video segments during the training phase and each video segment is an instance of a bag. 
They propose a deep MIL ranking loss as a hinge-loss formulation that also considers sparsity and temporal smoothness constraints. 
Anomalous and normal surveillance videos were partitioned into segments so that a video (bag) contains multiple segments (instances of a bag). 
To build and evaluate the proposed method, the authors consider the large-scale video anomaly dataset UCF-Crime which is composed of multiple anomalous events. From the obtained results, the proposed approach overcomes other anomaly detection state-of-the-art approaches in performance with 75.41\% and 1.9\% in terms of AUC and false alarm rate.

For training and evaluating our proposed multi-camera training strategy, we adopt this same MIL baseline framework as the backbone. This approach corresponds to a regression network trained under the MIL paradigm. In their study, the authors handle the anomaly detection task as a regression problem under MIL since only video labels are considered for model training. Their proposed solution takes a 3-layer fully connected neural network where the loss function contains restrictions of sparsity (anomaly scores are sparse since an anomaly usually occurs in a short time period) and smoothness (the anomaly score will vary smoothly). The MIL objective function is expressed as follows:

\begin{equation}
\begin{split}
\mathcal{L}(\mathcal{W}) = max(0, 1 - \max_{i \in \mathcal{B}_a} f(\mathcal{V}^i_a) + \max_{i \in \mathcal{B}_n} f(\mathcal{V}^i_n))  + \lambda_1\sum_i^{(n-1)}( f(\mathcal{V}^i_a) - f(\mathcal{V}^{i+1}_a)       )^2 + \lambda_2\sum_i^{n}f(\mathcal{V}^i_a) + \lambda_3 ||\mathcal{W}||_F
\end{split}
\end{equation}

Since it is expected that $f(\mathcal{V}^i_a) > f(\mathcal{V}^i_n)$ and the instance labels are unknown, the strategy consists of a MI ranking loss where $ \max_{i \in \mathcal{B}_a} f(\mathcal{V}^i_a) > \max_{i \in \mathcal{B}_n} f(\mathcal{V}^i_n)$, which is expressed in the \textit{hinge loss} formulation in the loss function equation. In the original paper, each video is divided into $32$ non-overlapping segments, where each segment is an instance of the bag. To a feature vector for the whole segment, the authors take the average of all clip features within each segment. During the training phase, they randomly select  a batch with $30$ negative and $30$ positive bags for loss computation and backpropagation. In the context of surveillance videos, data is highly imbalanced in weakly supervised anomaly detection. Differently from \cite{SultaniEtAl2018}, a characteristic of our multi-camera video anomaly detection preprocessed I3D dataset  is  that each video has a varying amount of clips.

\subsection{Proposed Framework}

How to use the synchronized information from multiple overlapped cameras for the anomaly detection task?  There is more than one way to combine them. 

A first way (1) is the direct concatenation of the preprocessed feature vectors $x = [x_1, x_2]$, where $x_1$ and $x_1$ are the feature vectors from camera 1 and 2 in the two-cameras case, respectively.  This immediate solution has the drawback of exponentially increasing the amount of labeled data needed to train predictive models. A second way (2) is the previous fusion of low level video attributes. From the concatenation of synchronized segments of each camera in a single joined segment, there are no growth in the dimension of the final I3D feature vector. Another possible way (3) is the training of separated models  per camera and the combination with a late fusion scheme as \cite{hao2020anomaly}. There are various published late fusion strategies in the literature. The performance of these approaches relies heavily on the engineering of the classifiers committee, and substantially increases time, design complexity, and model size. This work proposes an approach along another line: The conception of loss functions that simultaneously consider information from multiple cameras.

\begin{figure}[!ht]
    \centering
    \includegraphics[width=0.90\textwidth]{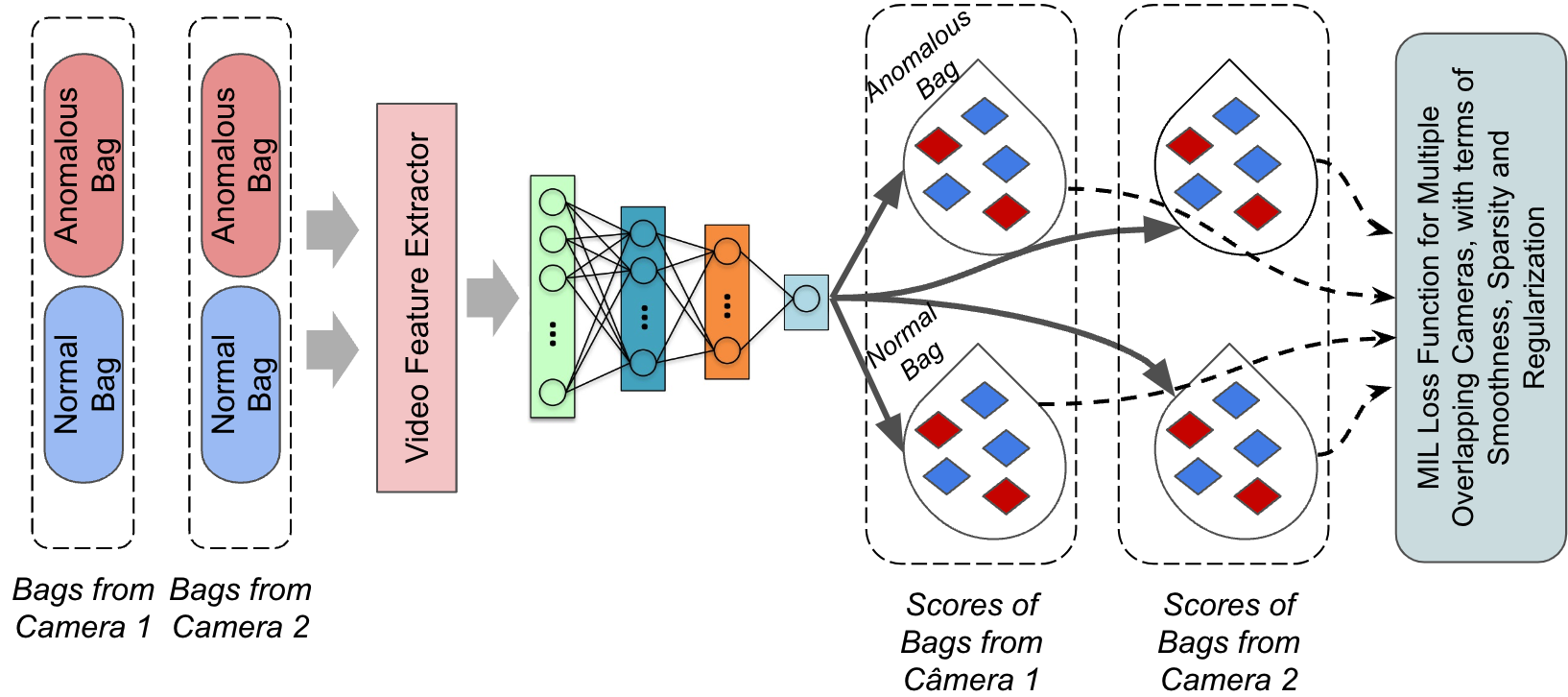}
    \caption{Operation of the Multiple Camera MIL (MC-MIL) framework inspired by Sultani et al. Bags of multiple simultaneous views generate bags of scores in the forward step. A loss function $L_{MC}(\mathcal{W})$, which combines scores from the views, is designed to train the regressive network in the backward step. In particular, setting $L_{MC}(\mathcal{W}) = max(L_{C_1}(\mathcal{W}), L_{C_2}(\mathcal{W})$ for two cameras, where $L_{C_i}(\mathcal{W})$ is the function of loss of Sultani et al. to camera $C_i$ was a good result.}
    \label{fig:McMilTheoreticalFramework}
\end{figure}

The proposed framework (4) is an extension of the \cite{SultaniEtAl2018} framework previously described for training with data from multiple cameras. We can insert more cameras to the processing pipeline by simply stacking them. We represent this scheme in Figure \ref{fig:McMilTheoreticalFramework}. Note that the regressive network is the same as  the authors, so there is no growth in model dimensions or training data required per camera. The intuition is the conception of loss functions that simultaneously consider information from multiple cameras.

For the backward step a proper loss function is computed over all bags of scores to compose the loss function, as represented by:

\begin{equation}
\mathcal{L}_{MC}(\mathcal{W}) = \mathcal{L}_{proper}(\{\bm{S}_a^{C_1}, \bm{S}_n^{C_1}, \bm{S}_a^{C_2}, \bm{S}_n^{C_2}\}) +   \mathcal{L}_{suavization} + \mathcal{L}_{esparsity} + \mathcal{L}_{regularization}
\end{equation}

\noindent
where $L_{proper}$ is defined over bags of scores and the other terms are as in the MIL loss function. Note that the $L_{proper}$ component leaves margin for the proposal of a wide range of loss functions. 
Next we present a concept proof which is an specialization of the proposed framework.

\subsection{Proof of Concept}
As a proof of concept, we specialize the framework of Section 2.3 for two cameras with $L_{MC}(\mathcal{W}) = Max(L_{c_1}(\mathcal{W}), L_{C_2}(\mathcal{W}))$, where $L_{C_i}(\mathcal{W})$ is the Sultani et al. loss function for camera $C_i$.
We trained the MIL approach based on the code available on this link\footnote{https://github.com/ekosman/AnomalyDetectionCVPR2018-Pytorch}, which provides a re-implementation of their approach in PyTorch\footnote{https://pytorch.org/} framework.
For the training of the regressor network, video bags contains a fixed number of $32$ non-overlapping segments, where each segment is an instance of the bag.  Differently from the authors, we consider that each video in our approach has a varying amount of clips, such as \cite{wan2020weakly}. Since the utilized preprocessed I3D data contains bags with a variable number of video clips, we then adapt the source code to allow the MIL training from video bags with a variable number of clips. It is important to note that our proposal does not increase the dimensions of the original regression network.

\begin{figure}[!ht]
    \centering
    \includegraphics[width=0.7\textwidth]{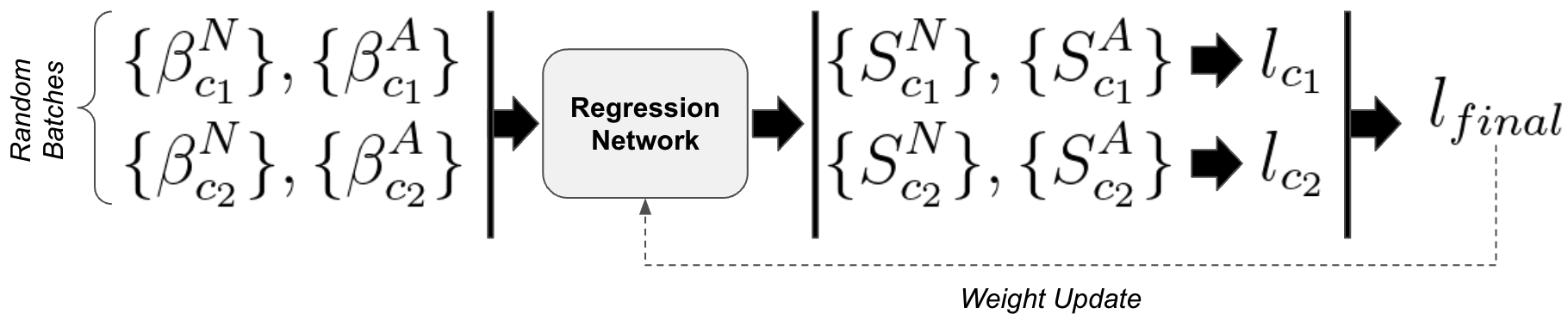}
    \caption{Multi-Camera Training Scheme with Sultani et al MIL Loss}
    \label{fig:MVLossFig}
\end{figure}

The following outlines our multi-camera training scheme: Let's consider two overlapped camera views 1 and 2. For each camera view $i$, where $i=\{1, 2\}$, the regression network receives a batch composed with a set of normal clip bags $\{ \beta^N_{c_i} \}$ and a set of abnormal ones, $\{ \beta^A_{c_i} \}$. The regression network outputs a set of normal clip scores $\{ S^N_{c_i} \}$ and a set of abnormal ones,
$\{ S^A_{c_i} \}$. Then, we obtain a loss $l_{c_{i}}$ from the computed clip scores. A combined loss function $l_{final}$ of $l_{c_{1}}$ and $l_{c_{2}}$ is used to adjust the network parameters. We evaluate loss combinations by minimum, maximum and mean. In our experiments, the loss combination by maximum achieves the best results. In the Figure \ref{fig:MVLossFig}, we present the proposed training scheme for the multi-camera video anomaly detection task.

As a result, the multi-camera regression network is able to produce an anomaly score for each of the cameras given a new unseen multi-camera clip. A late score fusion function can be used to combine the anomaly outputs yielded by each one of two camera view. We consider the following three late score fusion strategies to combine the anomaly scores yielded by regression network: linear combination (LC) ($S_{LC} \leftarrow \beta S_{C_{1}} + (1-\beta) S_{C_{2}}, \beta \in [0,1]$), maximum (Max) ($S_{MAX} \leftarrow \max(S_{C_{1}}, S_{C_{2}})$), and minimum (Min) ($S_{MIN} \leftarrow \min(S_{C_{1}}, S_{C_{2}})$). In our experiments, the score combination by maximum achieves the best results.

For the evaluation at the frame level, we describe the test segment partition by $\mathbf{X}_{test} = \{ (X_i, {yfs}_i) \}$, where ${yfs}_i$ is a sequence of  $16$ frame labels obtained from ground truth variable ${yf}_i$.  Since we build the generated models at the clip level, we mapped each predicted test clip output to the corresponding output frame sequence (each video clip corresponds to a sequence of $16$ frames) to allow evaluation at the frame level.

We consider six different performance metrics to report the achieved results. Let \textit{TP} be the number of true positives, \textit{FP} the number of false positives, \textit{TN} the number of true negatives, and \textit{FN} the number of false negatives. For each method and for each experiment, we computed the following $6$ metrics for performance evaluation: Area under Curve (AUC), False Positive Rate ($FPR=\frac{FP}{FP+TN}$) or False Alarm Rate (FAR), Balanced Accuracy ($BACC = \frac{TPR+TNR}{2}$), Precision ($PREC=\frac{TP}{TP+FP}$), Recall ($REC=\frac{TP}{TP+FN}$), and F1-score ($F1=2 \cdot \frac{PREC \cdot REC}{PREC + REC}$).  We reported FPR metric at $50\%$ threshold.

We compare our proposed multi-camera approach with the vanilla single-camera multiple instance approach (SC-MIL) with the MIL ranking loss function referred to in the previous section which was trained and evaluated for each one of the four cameras. Since we have four cameras in the video anomaly detection dataset, we evaluate the proposed multiple-camera learning scheme for all six combinations of two cameras: (C1,C2), (C1,C3), (C1,C4),(C2,C3),(C2,C4), and (C3,C4). To obtain a final prediction for a pair of cameras, we compute the score combination with the functions Linear Combination (LC), Maximum (Max), and Minimum (Min) of multiple camera anomaly output scores. In our experiments, the score combination by maximum achieves the best results.

\section{Results and Discussion}

This section presents the quantitative results of our proposed multi-camera multiple-instance (MC-MIL) approach for multi-camera video anomaly detection.

We summarize in Table \ref{tab:results-main} the best performance results for each pair of cameras and the late decision function of output scores. We also provide the baseline results for the single-camera MIL (SC-MIL) setting for further analysis and comparison. The performance evaluation of compared models considered ground truths at the frame level. As a training loss combination scheme, we only present the cost function combined with the maximum of the two loss functions in this Table, given that this combination was the one that generated better results in terms of the AUC metric. We repeat the single-camera baseline and multi-camera proposal $5$ times and the results are recorded for mean, standard deviation, minimum and maximum values.

\begin{table}[!h]
\centering
\begin{tabular}{ | c | c | c | c | c | c | c | c | }
\hline
\cellcolor[HTML]{FFFFFF}\textbf{Training} & \textbf{Test} & \textbf{AUC (\%)} & \textbf{FPR (\%)} & \textbf{BACC (\%)} & \textbf{PREC (\%)} & \textbf{REC (\%)} & \textbf{F1 (\%)} \\
\hline

\multirow{4}{*}{\textbf{SC-AMI}}                                          & C1    & \begin{tabular}[c]{@{}c@{}}91.6 ± 0.59\\  (90.92,92.5)\end{tabular}   & \begin{tabular}[c]{@{}c@{}}1.26 ± 0.18\\  (0.91,1.35)\end{tabular} & \begin{tabular}[c]{@{}c@{}}61.37 ± 1.24\\  (59.9,63.67)\end{tabular}  & \begin{tabular}[c]{@{}c@{}}77.81 ± 3.8\\  (74.48,85.27)\end{tabular}  & \begin{tabular}[c]{@{}c@{}}23.99 ± 2.32\\  (21.15,28.25)\end{tabular} & \begin{tabular}[c]{@{}c@{}}36.66 ± 3.12\\  (32.95,42.44)\end{tabular} \\
& C2    & \begin{tabular}[c]{@{}c@{}}95.39 ± 0.17\\  (95.12,95.65)\end{tabular} & \begin{tabular}[c]{@{}c@{}}0.91 ± 0.0\\  (0.91,0.91)\end{tabular}  & \begin{tabular}[c]{@{}c@{}}63.91 ± 0.48\\  (63.67,64.86)\end{tabular} & \begin{tabular}[c]{@{}c@{}}85.47 ± 0.39\\  (85.27,86.25)\end{tabular} & \begin{tabular}[c]{@{}c@{}}28.72 ± 0.95\\  (28.25,30.62)\end{tabular} & \begin{tabular}[c]{@{}c@{}}\textbf{42.99} ± 1.1\\  (42.44,45.2)\end{tabular}   \\
& C3    & \begin{tabular}[c]{@{}c@{}}93.75 ± 0.25\\  (93.38,94.16)\end{tabular} & \begin{tabular}[c]{@{}c@{}}1.06 ± 0.01\\  (1.05,1.07)\end{tabular} & \begin{tabular}[c]{@{}c@{}}62.71 ± 0.92\\  (61.96,64.33)\end{tabular} & \begin{tabular}[c]{@{}c@{}}82.21 ± 0.88\\  (81.25,83.75)\end{tabular} & \begin{tabular}[c]{@{}c@{}}26.48 ± 1.85\\  (25.0,29.73)\end{tabular}  & \begin{tabular}[c]{@{}c@{}}40.03 ± 2.2\\  (38.24,43.89)\end{tabular}  \\
& C4    & \begin{tabular}[c]{@{}c@{}}95.19 ± 0.27\\  (94.85,95.49)\end{tabular} & \begin{tabular}[c]{@{}c@{}}1.72 ± 0.1\\  (1.52,1.79)\end{tabular}  & \begin{tabular}[c]{@{}c@{}}59.66 ± 1.15\\  (58.59,61.74)\end{tabular} & \begin{tabular}[c]{@{}c@{}}69.26 ± 3.23\\  (66.67,75.45)\end{tabular} & \begin{tabular}[c]{@{}c@{}}21.03 ± 2.22\\  (18.93,25.0)\end{tabular}  & \begin{tabular}[c]{@{}c@{}}32.25 ± 2.95\\  (29.49,37.56)\end{tabular} \\

\hline
\multirow{3}{*}{\begin{tabular}[c]{@{}c@{}}C1C2\\  (MC-AMI)\end{tabular}} & C1    & \begin{tabular}[c]{@{}c@{}}90.26 ± 0.88\\  (89.06,91.75)\end{tabular} & \begin{tabular}[c]{@{}c@{}}0.91 ± 0.0\\  (0.91,0.91)\end{tabular}  & \begin{tabular}[c]{@{}c@{}}61.31 ± 0.0\\  (61.31,61.31)\end{tabular}  & \begin{tabular}[c]{@{}c@{}}82.81 ± 0.0\\  (82.81,82.81)\end{tabular}  & \begin{tabular}[c]{@{}c@{}}23.52 ± 0.0\\  (23.52,23.52)\end{tabular}  & \begin{tabular}[c]{@{}c@{}}36.64 ± 0.0\\  (36.64,36.64)\end{tabular}  \\
& C2    & \begin{tabular}[c]{@{}c@{}}95.89 ± 0.16\\  (95.68,96.09)\end{tabular} & \begin{tabular}[c]{@{}c@{}}0.91 ± 0.0\\  (0.91,0.91)\end{tabular}  & \begin{tabular}[c]{@{}c@{}}72.67 ± 1.61\\  (70.77,75.51)\end{tabular} & \begin{tabular}[c]{@{}c@{}}90.42 ± 0.58\\  (89.69,91.41)\end{tabular} & \begin{tabular}[c]{@{}c@{}}46.24 ± 3.21\\  (42.46,51.92)\end{tabular} & \begin{tabular}[c]{@{}c@{}}61.14 ± 2.91\\  (57.63,66.23)\end{tabular} \\
& C1,C2 & \begin{tabular}[c]{@{}c@{}}96.34 ± 0.28\\  (95.84,96.68)\end{tabular} & \begin{tabular}[c]{@{}c@{}}0.91 ± 0.0\\  (0.91,0.91)\end{tabular}  & \begin{tabular}[c]{@{}c@{}}73.61 ± 1.21\\  (71.96,75.51)\end{tabular} & \begin{tabular}[c]{@{}c@{}}90.77 ± 0.42\\  (90.18,91.41)\end{tabular} & \begin{tabular}[c]{@{}c@{}}48.14 ± 2.41\\  (44.82,51.92)\end{tabular} & \begin{tabular}[c]{@{}c@{}}\textbf{62.88} ± 2.16\\  (59.88,66.23)\end{tabular} \\

\hline
\multirow{3}{*}{\begin{tabular}[c]{@{}c@{}}C1C3\\  (MC-AMI)\end{tabular}} 
& C1    & \begin{tabular}[c]{@{}c@{}}92.36 ± 0.98\\  (90.68,93.61)\end{tabular} & \begin{tabular}[c]{@{}c@{}}0.73 ± 0.22\\  (0.47,0.91)\end{tabular} & \begin{tabular}[c]{@{}c@{}}61.4 ± 0.11\\  (61.31,61.53)\end{tabular}  & \begin{tabular}[c]{@{}c@{}}85.82 ± 3.69\\  (82.81,90.34)\end{tabular} & \begin{tabular}[c]{@{}c@{}}23.52 ± 0.0\\  (23.52,23.52)\end{tabular}  & \begin{tabular}[c]{@{}c@{}}36.91 ± 0.33\\  (36.64,37.32)\end{tabular} \\
& C3    & \begin{tabular}[c]{@{}c@{}}94.05 ± 0.32\\  (93.59,94.59)\end{tabular} & \begin{tabular}[c]{@{}c@{}}1.32 ± 0.22\\  (1.05,1.52)\end{tabular} & \begin{tabular}[c]{@{}c@{}}63.32 ± 0.79\\  (62.05,64.2)\end{tabular}  & \begin{tabular}[c]{@{}c@{}}79.94 ± 2.09\\  (77.5,83.04)\end{tabular}  & \begin{tabular}[c]{@{}c@{}}27.96 ± 1.74\\  (25.15,29.88)\end{tabular} & \begin{tabular}[c]{@{}c@{}}41.38 ± 1.79\\  (38.46,43.35)\end{tabular} \\
& C1,C3 & \begin{tabular}[c]{@{}c@{}}94.63 ± 0.21\\  (94.39,94.98)\end{tabular} & \begin{tabular}[c]{@{}c@{}}1.34 ± 0.22\\  (1.07,1.52)\end{tabular} & \begin{tabular}[c]{@{}c@{}}73.9 ± 0.54\\  (73.58,74.98)\end{tabular}  & \begin{tabular}[c]{@{}c@{}}87.26 ± 1.94\\  (85.68,89.84)\end{tabular} & \begin{tabular}[c]{@{}c@{}}49.14 ± 0.95\\  (48.67,51.04)\end{tabular} & \begin{tabular}[c]{@{}c@{}}\textbf{62.87} ± 1.17\\  (62.08,65.09)\end{tabular} \\

\hline
\multirow{3}{*}{\begin{tabular}[c]{@{}c@{}}C1C4\\  (MC-AMI)\end{tabular}} & C1    & \begin{tabular}[c]{@{}c@{}}94.31 ± 0.66\\  (93.71,95.23)\end{tabular} & \begin{tabular}[c]{@{}c@{}}0.96 ± 0.1\\  (0.91,1.16)\end{tabular}  & \begin{tabular}[c]{@{}c@{}}60.2 ± 0.16\\  (60.12,60.52)\end{tabular}  & \begin{tabular}[c]{@{}c@{}}80.62 ± 1.25\\  (78.12,81.25)\end{tabular} & \begin{tabular}[c]{@{}c@{}}21.36 ± 0.42\\  (21.15,22.19)\end{tabular} & \begin{tabular}[c]{@{}c@{}}33.77 ± 0.4\\  (33.57,34.56)\end{tabular}  \\
                                                                          & C4    & \begin{tabular}[c]{@{}c@{}}93.95 ± 0.2\\  (93.58,94.13)\end{tabular}  & \begin{tabular}[c]{@{}c@{}}2.01 ± 0.4\\  (1.32,2.4)\end{tabular}   & \begin{tabular}[c]{@{}c@{}}60.65 ± 1.06\\  (58.81,61.52)\end{tabular} & \begin{tabular}[c]{@{}c@{}}68.67 ± 3.28\\  (63.75,72.73)\end{tabular} & \begin{tabular}[c]{@{}c@{}}23.31 ± 2.38\\  (18.93,25.0)\end{tabular}  & \begin{tabular}[c]{@{}c@{}}34.71 ± 2.66\\  (30.05,36.9)\end{tabular}  \\
                                                                          & C1,C4 & \begin{tabular}[c]{@{}c@{}}94.81 ± 0.2\\  (94.49,95.09)\end{tabular}  & \begin{tabular}[c]{@{}c@{}}2.06 ± 0.39\\  (1.35,2.4)\end{tabular}  & \begin{tabular}[c]{@{}c@{}}60.95 ± 0.8\\  (59.9,61.92)\end{tabular}   & \begin{tabular}[c]{@{}c@{}}68.68 ± 3.69\\  (63.75,74.48)\end{tabular} & \begin{tabular}[c]{@{}c@{}}23.96 ± 1.8\\  (21.15,26.04)\end{tabular}  & \begin{tabular}[c]{@{}c@{}}\textbf{35.46} ± 1.93\\  (32.95,37.77)\end{tabular} \\
\hline
\multirow{3}{*}{\begin{tabular}[c]{@{}c@{}}C2C3\\  (MC-AMI)\end{tabular}} & C2    & \begin{tabular}[c]{@{}c@{}}96.15 ± 0.36\\  (95.46,96.46)\end{tabular} & \begin{tabular}[c]{@{}c@{}}0.73 ± 0.22\\  (0.47,0.91)\end{tabular} & \begin{tabular}[c]{@{}c@{}}63.52 ± 0.92\\  (62.49,65.08)\end{tabular} & \begin{tabular}[c]{@{}c@{}}87.65 ± 3.42\\  (84.13,92.41)\end{tabular} & \begin{tabular}[c]{@{}c@{}}27.78 ± 1.77\\  (25.89,30.62)\end{tabular} & \begin{tabular}[c]{@{}c@{}}42.16 ± 2.23\\  (39.59,46.0)\end{tabular}  \\
                                                                          & C3    & \begin{tabular}[c]{@{}c@{}}93.58 ± 0.16\\  (93.27,93.72)\end{tabular} & \begin{tabular}[c]{@{}c@{}}1.25 ± 0.22\\  (1.07,1.52)\end{tabular} & \begin{tabular}[c]{@{}c@{}}63.06 ± 0.11\\  (62.93,63.15)\end{tabular} & \begin{tabular}[c]{@{}c@{}}80.39 ± 2.7\\  (77.08,82.59)\end{tabular}  & \begin{tabular}[c]{@{}c@{}}27.37 ± 0.0\\  (27.37,27.37)\end{tabular}  & \begin{tabular}[c]{@{}c@{}}40.82 ± 0.35\\  (40.39,41.11)\end{tabular} \\
                                                                          & C2,C3 & \begin{tabular}[c]{@{}c@{}}95.33 ± 0.19\\  (95.09,95.55)\end{tabular} & \begin{tabular}[c]{@{}c@{}}1.25 ± 0.22\\  (1.07,1.52)\end{tabular} & \begin{tabular}[c]{@{}c@{}}74.66 ± 0.87\\  (73.58,75.94)\end{tabular} & \begin{tabular}[c]{@{}c@{}}88.31 ± 1.74\\  (85.68,89.84)\end{tabular} & \begin{tabular}[c]{@{}c@{}}50.56 ± 1.77\\  (48.67,53.4)\end{tabular}  & \begin{tabular}[c]{@{}c@{}}\textbf{64.28} ± 1.49\\  (62.08,66.12)\end{tabular} \\
\hline
\multirow{3}{*}{\begin{tabular}[c]{@{}c@{}}C2C4\\  (MC-AMI)\end{tabular}} & C2    & \begin{tabular}[c]{@{}c@{}}95.99 ± 0.29\\  (95.64,96.45)\end{tabular} & \begin{tabular}[c]{@{}c@{}}0.91 ± 0.0\\  (0.91,0.91)\end{tabular}  & \begin{tabular}[c]{@{}c@{}}61.54 ± 0.89\\  (60.12,62.49)\end{tabular} & \begin{tabular}[c]{@{}c@{}}83.03 ± 1.07\\  (81.25,84.13)\end{tabular} & \begin{tabular}[c]{@{}c@{}}23.99 ± 1.77\\  (21.15,25.89)\end{tabular} & \begin{tabular}[c]{@{}c@{}}37.21 ± 2.25\\  (33.57,39.59)\end{tabular} \\
                                                                          & C3    & \begin{tabular}[c]{@{}c@{}}94.23 ± 0.14\\  (94.0,94.38)\end{tabular}  & \begin{tabular}[c]{@{}c@{}}1.77 ± 0.36\\  (1.35,2.4)\end{tabular}  & \begin{tabular}[c]{@{}c@{}}60.46 ± 0.75\\  (59.68,61.74)\end{tabular} & \begin{tabular}[c]{@{}c@{}}70.68 ± 4.22\\  (63.75,75.45)\end{tabular} & \begin{tabular}[c]{@{}c@{}}22.69 ± 1.47\\  (21.15,25.0)\end{tabular}  & \begin{tabular}[c]{@{}c@{}}34.32 ± 1.9\\  (32.35,37.56)\end{tabular}  \\
                                                                          & C2,C4 & \begin{tabular}[c]{@{}c@{}}96.2 ± 0.23\\  (95.89,96.6)\end{tabular}   & \begin{tabular}[c]{@{}c@{}}1.77 ± 0.36\\  (1.35,2.4)\end{tabular}  & \begin{tabular}[c]{@{}c@{}}64.72 ± 1.36\\  (63.23,66.48)\end{tabular} & \begin{tabular}[c]{@{}c@{}}76.76 ± 3.43\\  (72.81,80.9)\end{tabular}  & \begin{tabular}[c]{@{}c@{}}31.21 ± 2.8\\  (28.25,34.47)\end{tabular}  & \begin{tabular}[c]{@{}c@{}}\textbf{44.31} ± 2.98\\  (40.99,48.34)\end{tabular} \\
\hline
\multirow{3}{*}{\begin{tabular}[c]{@{}c@{}}C3C4\\  (MC-AMI)\end{tabular}} & C3    & \begin{tabular}[c]{@{}c@{}}94.69 ± 0.44\\  (94.26,95.39)\end{tabular} & \begin{tabular}[c]{@{}c@{}}1.07 ± 0.0\\  (1.07,1.07)\end{tabular}  & \begin{tabular}[c]{@{}c@{}}63.86 ± 1.21\\  (61.96,65.51)\end{tabular} & \begin{tabular}[c]{@{}c@{}}83.22 ± 1.2\\  (81.25,84.77)\end{tabular}  & \begin{tabular}[c]{@{}c@{}}28.79 ± 2.41\\  (25.0,32.1)\end{tabular}   & \begin{tabular}[c]{@{}c@{}}42.74 ± 2.84\\  (38.24,46.57)\end{tabular} \\
                                                                          & C4    & \begin{tabular}[c]{@{}c@{}}93.71 ± 0.23\\  (93.38,94.05)\end{tabular} & \begin{tabular}[c]{@{}c@{}}1.49 ± 0.07\\  (1.35,1.52)\end{tabular} & \begin{tabular}[c]{@{}c@{}}60.43 ± 0.26\\  (59.9,60.56)\end{tabular}  & \begin{tabular}[c]{@{}c@{}}73.74 ± 0.37\\  (73.56,74.48)\end{tabular} & \begin{tabular}[c]{@{}c@{}}22.33 ± 0.59\\  (21.15,22.63)\end{tabular} & \begin{tabular}[c]{@{}c@{}}34.29 ± 0.67\\  (32.95,34.62)\end{tabular} \\
                                                                          & C3,C4 & \begin{tabular}[c]{@{}c@{}}95.39 ± 0.24\\  (94.99,95.69)\end{tabular} & \begin{tabular}[c]{@{}c@{}}1.52 ± 0.0\\  (1.52,1.52)\end{tabular}  & \begin{tabular}[c]{@{}c@{}}71.92 ± 0.58\\  (71.21,72.39)\end{tabular} & \begin{tabular}[c]{@{}c@{}}84.78 ± 0.33\\  (84.38,85.05)\end{tabular} & \begin{tabular}[c]{@{}c@{}}45.35 ± 1.16\\  (43.93,46.3)\end{tabular}  & \begin{tabular}[c]{@{}c@{}}\textbf{59.09} ± 1.07\\  (57.78,59.96\end{tabular} \\
\hline
\end{tabular}
\vspace{0.2cm}
\caption{Result of the performance evaluation of the MC-MIL algorithm for the different combinations of two cameras against the single camera configuration, on the re-labeled PETS-2009 dataset. Hidden layers contains a dropout rate of 60\% and 20000 iterations. Each iteration is a randomly chosen mini-batch of 30 normal + 30 anomalous bags, respectively.}
\label{tab:results-main}
\end{table}

When we consider only one camera in the inference phase, i.e., either cameras C1, C2, C3, or C4 in SC-MIL or MC-MIL training, we observe that the proposed MC-MIL scheme outperforms three of the four single-camera baselines: For the first camera, there is a performance gain 
in terms of AUC, respectively. 
In terms of F1-Score, we can also observe gains when we compare SC-MIL and MC-MIL results for only one camera in the inference phase. 

But it is when we fuse the decision scores of the two cameras by the maximum decision that the harmonic mean between precision and recall becomes more expressive in five of the four pairs of camera combinations. When we pay attention to the decision inference in MC-MIL for two cameras, we note a positive difference 
for the pairs (C1,C2), (C3, C4), (C1,C3), (C2,C3), and (C2,C4), respectively, although the pair (C1,C4) has no performance gain in F1-Score.

\begin{figure}[!ht]
    \centering
    \includegraphics[scale=0.60]{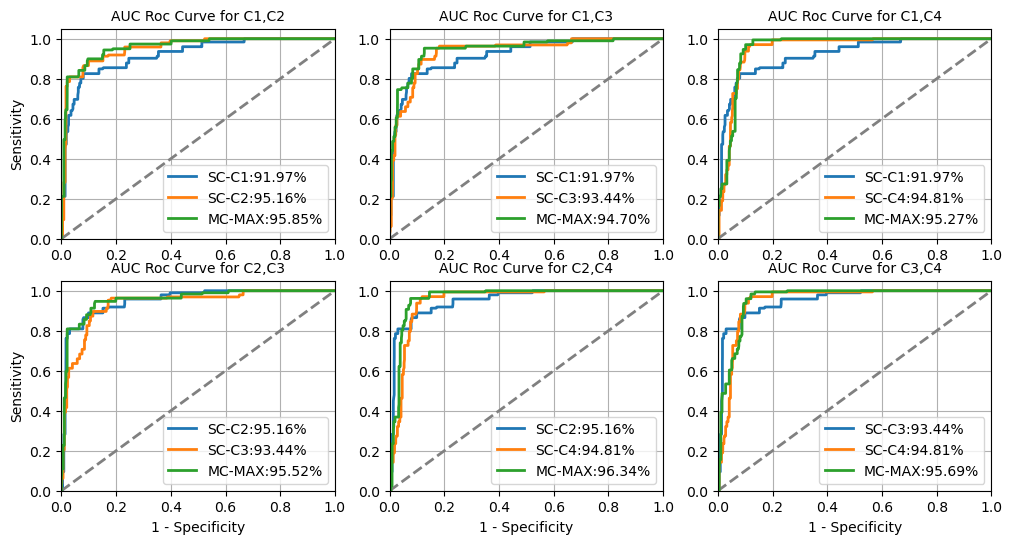}
    \caption{ROC Curves for Single-Camera and Multi-Camera MIL Approaches Trained and Evaluated with RGB I3D Features.}
    \label{fig:roc-curve}
\end{figure}

In Figure \ref{fig:roc-curve} we present the frame-level AUC ROC curves for our proposed approach MC-MIL in comparison with the baseline SC-MIL models for each pair of cameras. 
Although we can observe that, for all pairs of cameras, the curves of the SC-MIL and MC-MIL approaches seem similar and overlap in many regions of the graph of each camera combination,  for the pairs C1-C3 and C2-C4, there are still some regions of the graphic where the MC-MIL approach outperforms, even if by a marginal gain, the SC-MIL approach. Despite the ROC curves being close, we can verify subtle differences for F1-Score and AUC metrics, indicating the slightly superior performance of the MC-MIL approach over SC-MIL.
These results provide an indication of the greater robustness of the MC-MIL approach in terms of AUC and F1-Score for the context of using a single camera in the inference phase. Also, in the context when we have multiple cameras to support the decision inference, the simple late fusion by maximum is able to improve the tradeoff between precision and recall.  

\subsection{Qualitative Analysis}

\begin{figure}
    \centering
    \includegraphics[width=\linewidth]{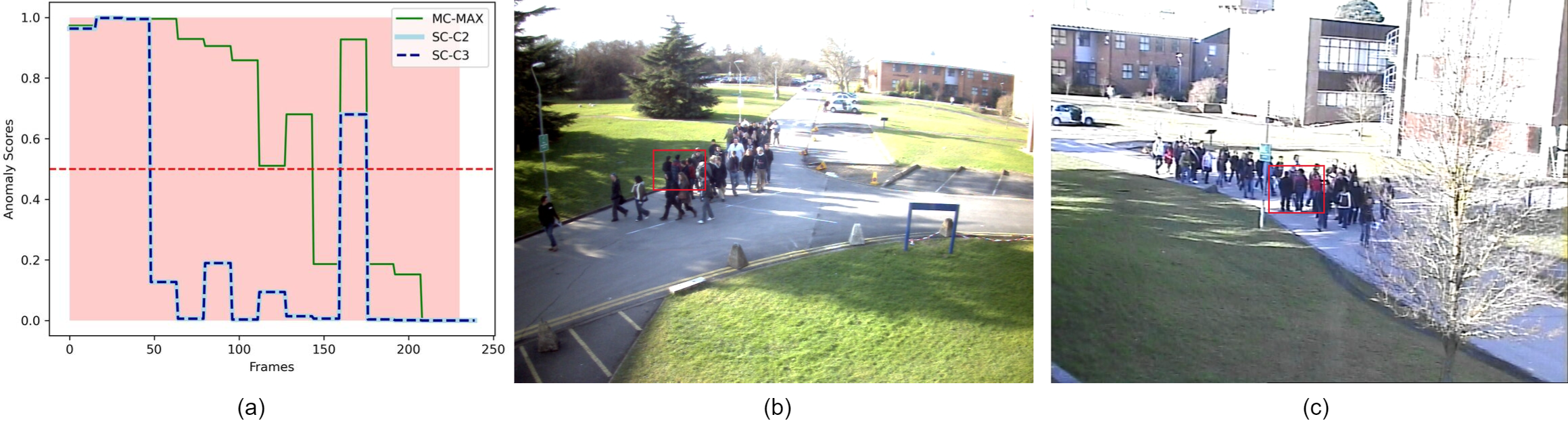}
    \caption{Visualization of Proposed Multi-Camera Approach and Single-Camera Baseline for an Anomalous Test Scene: (a) The Anomaly Scores throughout the video (b) Frame $100$ of Camera 2. (c) Frame $100$ of Camera 3.}
    \label{fig:qualitative-analysis}
\end{figure}

To further exemplify the performance of our proposed multi-camera approach against the single-camera baseline, we visualize the temporal predictions of the models for the anomaly test scene \textit{S3-Multiple-Flows--Time-14-13} for the camera pair (C2, C3), as showed in Figure \ref{fig:qualitative-analysis}. We present the analysis of our proposed approach in comparison with the single-camera baselines.  

In the video \textit{S3-Multiple-Flows--Time-14-13}, pedestrians standing in the middle of the intersection are considered an anomaly. The video begins with a small people crowd walking down the street. In front of them, four other people that are positioned immobile and lined up side by side in part of the traffic intersection. These three people remain motionless throughout the entire video, which is considered an anomaly here. When the flow approaches, this main flow goes around the three people without bump into them. 
The video is challenging since their spatial and motion patterns are very approximate to the normal videos. This event  is similar to a normal scene where people in crowd just walk. We mark with red bounding boxes the position of the three people remaining motionless. At the frame $100$, it is no longer possible to clearly distinguish the crowd and the people standing in the middle of the street. 
We can observe that the single-camera models performed poorly, while the multi-camera approach was able to detect the anomaly event even when the two groups of people were very close together.

 \section{New Results on Bag Scores Union and Multiview Network Architecture}

We also present results for another variation of our proposed multicamera multiple-instance framework. We define an strategy to combine the normal and abnormal bags of scores before passing them to the Mil loss function. The intuition is efficiently combine the information at bag level of both cameras for the network learning: during the training phase, we obtain a single anomaly score bag as $S_a^* \equiv mean (S_a^{C_1}, S_a^{C_2})$ from the respective bags of each camera. Then, we obtain each normal bag as $S_n^* \equiv mean (\beta_n^{C_1}, S_n^{C_2})$. Finally, the backpropagation is done through the loss function $\mathcal{L}_{MC}(\mathcal{W}) = \mathcal{L}_{Sultani} (\beta_a^*, \beta_n^*)$ of Sultani. 


\begin{equation}
\mathcal{L}_{MC}(\mathcal{W}) = \mathcal{L}_{Sultani}(S_a^*, S_n^*)
\end{equation}

We also make the assumption that the use of a single fully connected network to learn the characteristics of both cameras can limit the model capacity in learning the particularities of each individual cameras, since both cameras will share the same network weights. From the inspiration in deep multiview learning (\cite{Wang2022Multiview}), we also equip all proposed framework with a multiview network to learn the specific properties of each individual camera.  

\begin{figure}[!ht]
    \centering
    \includegraphics[width=0.90\textwidth]{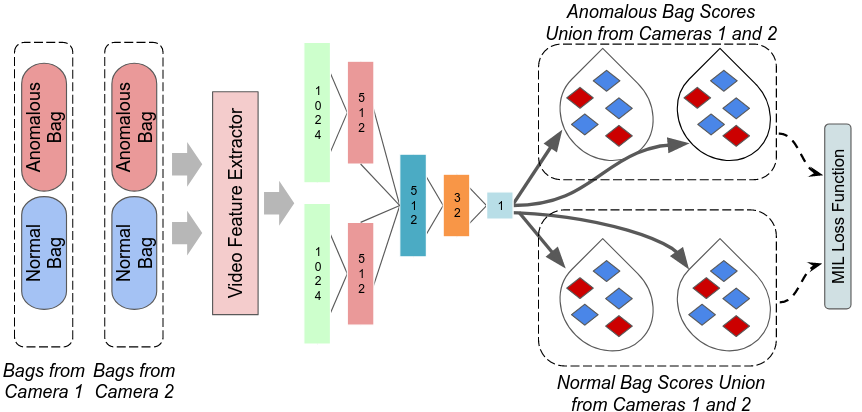}
    \caption{Operation of the Multiple Camera MIL (MC-MIL) framework with the multiview neural network and the combination of scores of normal and anomalous bags. }
    \label{fig:MtvArchMcMilTheoreticalFramework}
\end{figure}

Since The regressive network of Sultani contains  two hidden layers  with $512$ and $32$ neurons and an single-neuron output layer, we modified the original network so new weight matrices  will exist for each camera as described in Figure \ref{fig:MtvArchMcMilTheoreticalFramework}.  
We present the achieved results in Table 3.

\begin{table}[!h]
\centering
\begin{tabular}{|l|c|c|c|}
\hline
\textbf{Approach} & \textbf{AUC (\%)} & \textbf{FPR (\%)} & \textbf{F1 (\%)}\\
\hline
Supervised -- LGBM (C2)   & 98.03  & 0.91 & 79.01\\
\hline
Supervised -- LGBM (C3)  & 92.6  & 4.24  & 67.41 \\
\hline
SC-MIL (C2) & \begin{tabular}[c]{@{}c@{}}95.39 ± 0.17\\ (95.12, 95.65)\end{tabular} & \begin{tabular}[c]{@{}c@{}}0.91 ± 0.0\\  (0.91,0.91) \end{tabular} & \begin{tabular}[c]{@{}c@{}}42.99 ± 1.1\\  (42.44,45.2)\end{tabular} \\
\hline
SC-MIL (C3) & \begin{tabular}[c]{@{}c@{}}93.75 ± 0.25\\ (93.38,94.16)\end{tabular} & \begin{tabular}[c]{@{}c@{}}1.06 ± 0.01\\  (1.05,1.07)\end{tabular} & \begin{tabular}[c]{@{}c@{}}40.03 ± 2.2\\  (38.24,43.89)\end{tabular} \\
\hline
MC-MIL Approach (C2,C3) & \begin{tabular}[c]{@{}c@{}}95.33 ± 0.19\\  (95.09,95.55)\end{tabular} & \begin{tabular}[c]{@{}c@{}}1.25 ± 0.22\\  (1.07,1.52)\end{tabular} & \begin{tabular}[c]{@{}c@{}}64.28 ± 1.49\\  (62.08,66.12)\end{tabular} \\
\hline
Score Late Fusion (2,3) By Max          & \begin{tabular}[c]{@{}c@{}}95.22 ± 0.19\\  (94.99,95.53)\end{tabular} & \begin{tabular}[c]{@{}c@{}}1.07 ± 0.0\\  (1.07,1.07)\end{tabular}  & \begin{tabular}[c]{@{}c@{}}65.08 ± 1.29\\  (63.03,67.1)\end{tabular}  \\
\hline
New Approach (Bag Union + Multiview Network) & \begin{tabular}[c]{@{}c@{}}95.14 ± 0.34\\  (94.66,95.53)\end{tabular} & \begin{tabular}[c]{@{}c@{}}1.52 ± 0.28\\  (1.07,1.96)\end{tabular} & \begin{tabular}[c]{@{}c@{}}\textbf{74.89 ± 1.35}\\  \textbf{(73.53,76.94)}\end{tabular} \\
\hline
\end{tabular}
\vspace{0.2cm}
\caption{Result of the performance evaluation of the MC-MIL algorithm for the cameras C2 and C3 (Pets2009 Dataset)}
\end{table}

The single-camera supervised baseline allows us to know the maximum performance that could be obtained when one of the cameras is used and how far or close the compared approaches are from this target. We use LightGBM\cite{ke2017lightgbm} as the supervised baseline, which is an approach for iterative and parallel training with weak classifiers (decision trees). This method allows parallel optimization and is known for fast training speed, high accuracy, memory consumption, and support for distributed processing (\cite{yi2020terrorist}). We also evaluate the late fusion by maximum of scores of each camera  using the algorithm proposed by \cite{SultaniEtAl2018} as the MIL backbone.

We can observe a significative performance gain when we use the strategy of bag score combination during the training of the network. This may indicate this approach helps highlight the more appropriate score values for the clips in bags, which improves network generalization. When we use the multiview network, the performance also improves, which holds the intuition of keeping shared and specific information at the same time in the neural network.

We also preprocess and adapt the multi-class Up-Fall Detection dataset \cite{martinez2019up} for the video anomaly detection task by extracting the I3D features from optical-flow video clips of the two overlapped camera video scenes. We considered the falling-related categories as anomalous events and the remaining categories as normal events. There are 559 video scenes (243 videos with anomalies and 316 normal ones), where each scene contains two overlapped camera views. 

\begin{table}[!h]
\centering
\begin{tabular}{|l|c|c|c|}
\hline
\textbf{Approach} & \textbf{AUC (\%)} & \textbf{FPR (\%)} & \textbf{F1 (\%)}\\
\hline
Supervised - LGBM (C1) & 99.71  & 01.04  & 84.77    \\
\hline
Supervised LGBM (C2) & 99.59 & 01.04   & 79.62 \\
\hline
McMiVad C1,C2   & \begin{tabular}[c]{@{}c@{}}97.53 ± 1.75 \\ (94.23,98.9)\end{tabular}  & \begin{tabular}[c]{@{}c@{}}20.61 ± 39.69 \\ (0.71,100.0)\end{tabular} & \begin{tabular}[c]{@{}c@{}}64.78 ± 29.59 \\ (5.64,81.38)\end{tabular} \\
\hline
Score Late Fusion (1,2) By Max                                                & \begin{tabular}[c]{@{}c@{}}99.02 ± 0.22 \\ (98.63,99.21)\end{tabular} & \begin{tabular}[c]{@{}c@{}}0.58 ± 0.02 \\ (0.55,0.6)\end{tabular}     & \begin{tabular}[c]{@{}c@{}}82.78 ± 0.43 \\ (82.38,83.59)\end{tabular} \\
\hline
New Approach (Bag Union + Multiview Network)                                  & \begin{tabular}[c]{@{}c@{}}98.13 ± 0.28 \\ (97.65,98.5)\end{tabular}  & \begin{tabular}[c]{@{}c@{}}0.55 ± 0.01 \\ (0.53,0.56)\end{tabular}    & \begin{tabular}[c]{@{}c@{}}\textbf{83.25 ± 0.46} \\ \textbf{(82.37,83.69)}\end{tabular} \\
\hline
\end{tabular}
\vspace{0.2cm}
\caption{Result of the performance evaluation of the MC-MIL algorithm for the cameras C2 and C3 (Up-Fall Dataset)}
\end{table}

We could also observe the robustness of the combination of multiview network and bag union strategy in the Up Fall dataset. We observe gains in terms of F1-Score in comparison with the default MC-MIL approach and the simple late score fusion strategy.

\section{Related Work}

There is a considerable number of research studies on video surveillance and video anomaly detection. Comprehensive reviews that expand the coverage of this section can be found in \cite{Patrikar2022ReviewVSurv}. This section overviews related studies about binary classification and multiple instance learning in video anomaly detection.

\nocite{Pereira2022Bracis22}
\cite{Pereira2022Bracis22} presents a wrapper-based multiple instance learning approach for video anomaly detection that applies a LightGBM model built with publicly available deep features constructed  with a clip-based instance generation strategy. They evaluate the approach with the single-camera ShanghaiTech dataset. 
To mitigate the redundant information in highly correlated deep features in the dataset the authors consider the removal of the higher correlated features from the computation and analysis of Pearson correlation matrix of training data at clip level. They compare the results against other commonly used methods and with the state-of-the-art literature. The authors observe that the proposed approach is able to overcome the frame-level results of the literature in terms of the AUC metric, although our technique suffers from a high false positive rate.

\nocite{hao2020anomaly}
In \cite{hao2020anomaly} the authors approach the task of video anomaly detection from a model based on two \textit{streams} to deal with RGB and optical flow data with specific neural networks for each information modality. The final anomaly score is made from the weighted fusion of the anomaly scores generated by each network. The intention is that the information from each stream can complement each other and thus improve the performance of the final detection. The task is treated as a regression problem under the multi-instance paradigm. For the training of each network, the authors make use of the MIL ranking cost function proposed by \cite{SultaniEtAl2018}. The authors also evaluate the performance impact regarding the use of the same or different number of layers in the networks. The authors observed that the merging of \textit{streams} with a different number of layers obtained better results than the use of networks with the same number of layers. The authors argue that merging in this way not only allows to use of the complementary information of the two \textit{streams} but also makes it possible to explore multi-scale information.

\cite{Putra2022Deep} present a deep neural network model to recognize human activities from multiple cameras by utilizing raw images to feed the network.The approach includes a feature extractor and a discriminator in order to capture the local and temporal information in data. The approach contains three components: a convolutional neural network to extract spatial information, an LSTM network to decode temporal information (MSLSTMRes - Multiple Stacked Long Short-term Memory Residual), and a dense layer with softmax activation to classify the feature patterns of all views. The approach performs a late data fusion by the concatenation of the processed data through network blocks CNN and LSTM for each camera before  delivering the information to the dense layer. In this sense, the final layer seeks to relate the information of multiple cameras to categorize the action. The approach has five input units, corresponding to the images of each camera. The authors also employ an attention mechanism to get information about the area of a moving object.

In this work, we evaluate the application of a multi-camera multiple-instance strategy to optimize the robustness of predictive models by combining loss function values of each camera view in an aggregated loss function for backpropagation and weight adjustment. The intuition of our proposal is that with the employment of a combined loss function that considers multiple camera views rather than conventional MIL training with one single camera, we will be able to improve the final capacity of the predictive model. From a multi-camera video anomaly dataset composed from the benchmark PETS 2009 dataset preprocessed as inflated 3D (I3D) clips in a fine-grained clip-based manner, we get promising results in comparison to the single-camera scenario.

\section{Conclusion}

In this work, we explore the anomaly detection problem in surveillance video by combining Multiple Instance Learning (MIL) with Multiple Camera Views (MC).
We are the first to propose the multiple-camera multiple-instance training scheme for the MIL algorithm of Sultani et al. [2018] which uses a combined loss function to take into account the multiple camera views of the same scene during the network weight adjustment. Our proposal does not increase the dimensions of the original regression network.
Due to the lack of a suitable dataset available, the PETS-2009 dataset has been re-labeled to perform proof-of-concept testing. The result shows a significant performance improvement in F1 score compared to the single-camera configuration.

The use of multiple camera views of the same scene opens up an avenue of research opportunities for improving the performance of anomaly detection in surveillance video. A number of loss functions derived from the original score rank function of Sultani et al. [2018] was published after them (\cite{FengEtAl2021MIST}, \cite{zhang2019temporal}, \cite{LiEtAl2022MultiviewVAD}). So, in addition to investigating the conception of a new multiple-camera multiple-instance loss function, we are investigating multitask and experimenting to apply these other previously published loss functions for the setting of multiple-camera multiple-instance learning.

\bibliographystyle{unsrtnat}
\bibliography{references}
\end{document}